\documentclass[10pt,twocolumn,letterpaper]{article}

\usepackage{wacv}      

\usepackage{graphicx}
\graphicspath{{../_ECCV_26__VETO/images/}}
\usepackage{booktabs}
\usepackage[accsupp]{axessibility}
\usepackage{algorithm}
\usepackage{algorithmic}
\usepackage{multirow}
\usepackage[table]{xcolor}
\definecolor{myblue}{RGB}{219, 234, 254}

\makeatletter
\renewcommand\paragraph{\@startsection{paragraph}{4}{\z@}%
  {0.5ex \@plus 0.2ex \@minus .2ex}%
  {-1em}%
  {\normalfont\normalsize\bfseries}}
\makeatother

\def\MethodName{VETO}
\def\MethodNameExt{Video Efficient Token Optimization}

\definecolor{wacvblue}{rgb}{0.21,0.49,0.74}
\usepackage[pagebackref,breaklinks,colorlinks,allcolors=wacvblue]{hyperref}
\usepackage[most]{tcolorbox}
\tcbuselibrary{skins}
\newtcolorbox{findingbox}[1]{%
  enhanced, rounded corners,
  colback=blue!4!white, colframe=blue!45!black,
  leftrule=2pt, rightrule=0.2pt, toprule=0.2pt, bottomrule=0.2pt,
  left=3pt, right=3pt, top=3pt, bottom=3pt,
  fonttitle=\bfseries\small, title={Finding #1},
  before skip=6pt, after skip=6pt,
}

\def\wacvPaperID{1189} 
\def\confName{WACV}
\def\confYear{2027}

\title{\MethodName: \MethodNameExt~for Vision-Language Models}

\author{
Gueter Josmy Faure$^{1}$ \quad
Hao Ping Wang$^{1}$ \quad
Min-Hung Chen$^{2}$ \quad
Winston H. Hsu$^{1}$\\ \\
$^{1}$National Taiwan University \qquad
$^{2}$NVIDIA
}

\begin{document}

\twocolumn[{%
	\renewcommand\twocolumn[1][]{#1}%
	\maketitle
	\begin{center}
		\newcommand{\teaserwidth}{\textwidth}
		\centerline{
			\includegraphics[width=\teaserwidth,clip]{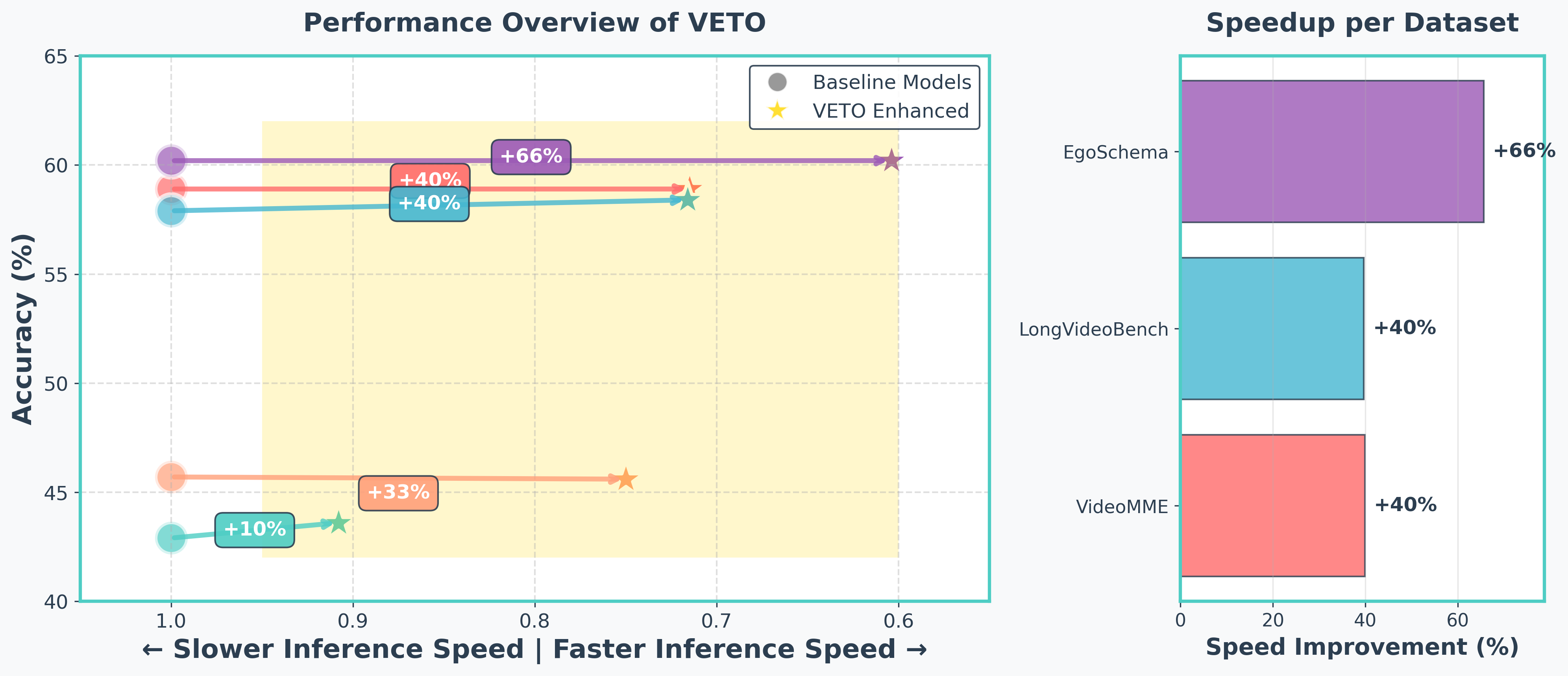}
		}
		\captionof{figure}{VETO performance overview. Our dual-axis compression places VETO in the efficiency–accuracy sweet spot: it reduces inference latency across model scales and benchmarks while preserving or improving task accuracy.}
		\vspace{-0.05in}
		\label{fig:teaser}
	\end{center}%
}]

\begin{abstract}
Processing long videos with Vision-Language Models (VLMs) is bottlenecked by
the quadratic cost of visual tokens, making long-form inference prohibitively
expensive.
While single-axis compression methods mitigate this, they hit a hard efficiency
floor because they treat spatial and temporal redundancy independently.
We present \textbf{VETO} (\MethodNameExt), a \emph{training-optional} plug-in
that eliminates this bottleneck through dual-axis compression: (i) an
\textbf{intra-frame} compressor that merges semantically similar tokens within
each frame via optimal-transport inspired matching, and (ii) an
\textbf{inter-frame} compressor that identifies and merges temporally redundant
frames.
The key design insight is hierarchical ordering: by first compressing spatial
dimensions, VETO drastically reduces the cost of subsequent global temporal
matching—bypassing the efficiency wall of single-axis approaches, with an
advantage that \emph{grows} with modern fully-fused attention infrastructure.
Empirically, VETO achieves up to \textbf{45\%} faster inference (e.g., on
LLaVA-OneVision-7B) while preserving or improving accuracy.
Under extreme token starvation (10\% budget), VETO outperforms VFlowOpt \cite{yang2025vflowopt}
(54.9\%), VisionZip \cite{yang2025visionzip} (52.6\%), and FastV \cite{chen2024image} (47.9\%) with \textbf{55.7\%} accuracy.
We demonstrate universal applicability across LLaVA-OneVision, InternVL-2.5,
and LongVA, with zero-shot accuracy preserved or improved in all cases.
\end{abstract}

\section{Introduction}
\label{sec:intro}

Vision-Language Models (VLMs) have transformed multimodal reasoning, enabling
systems to answer questions, summarize scenes, and follow instructions grounded
in visual
input~\cite{instructblip,liu2023visual,bai2023qwenvlversatilevisionlanguagemodel}.
Yet scaling these systems to long-form video remains a critical bottleneck:
attention costs grow quadratically with the total number of visual tokens.
For a video with $T$ frames and $N$ tokens per frame, the complexity is
$O\!\big((TN)^2\big)$, meaning multi-minute videos quickly produce tens of
thousands of tokens, rendering inference prohibitively expensive.

Existing solutions trade off accessibility and performance.
Training-based methods require substantial compute.
Training-free plug-ins (e.g., token pruning, merging, or frame
selection)~\cite{bolya2022token,chen2024image,faure2025hermes} are immediately
deployable, yet they overwhelmingly treat spatial and temporal redundancy
independently.
This single-axis paradigm harbors a critical flaw: relying solely on
inter-frame compression requires computing similarity over \emph{fully
uncompressed} spatial tokens, inflating latency.
Single-axis methods therefore hit a hard efficiency floor, preventing optimal
speed-accuracy trade-offs.

We introduce \textbf{VETO} (\MethodNameExt), a training-optional plug-in that
resolves this through a \emph{dual-axis} strategy.
VETO applies (i) an \emph{Intra-Frame Compressor} that merges semantically
similar tokens within each frame using optimal-transport-inspired assignment,
followed by (ii) an \emph{Inter-Frame Compressor} that globally identifies and
merges temporally redundant frames.
By executing spatial compression \emph{before} temporal matching, VETO slashes
the feature dimensions required for global similarity calculations, mitigating
the latency bottleneck of single-axis methods.
This reduces effective attention complexity to
$O\!\big(TN^2/k + T^2C\big)$ (with $C \ll N$), enabling substantial runtime
savings.

A critical property of VETO's design is that its advantage over single-axis
baselines is \emph{hardware-agnostic}—and in fact \emph{grows} on modern,
optimized attention infrastructure.
Fully-fused attention kernels effectively absorb per-frame spatial reductions
from intra-only methods, but cannot absorb VETO's total-sequence-length
reduction.
This means VETO's relative speedup margin over single-axis baselines increases
on modern deployments, making it an increasingly compelling choice as inference
stacks continue to be optimized.

\textbf{System contribution.}
Beyond the algorithmic design, VETO is immediately deployable as a drop-in
module requiring no custom CUDA kernels or architectural modifications.
It operates as a pure tensor transformation—applied between the visual encoder
and the LLM—and is fully compatible with standard attention implementations
including FlashAttention.
The end-to-end latency measurements reported in this paper (visual encoder +
token compression + LLM inference) constitute a realistic systems-level
evaluation of deployment cost.

We validate VETO across three VLM families: LLaVA-OneVision~\cite{li2024llava}
(0.5B and 7B), InternVL-2.5~\cite{chen2024expanding} (1B), and
LongVA~\cite{zhang2024long} (7B).
VETO consistently reduces end-to-end latency while preserving or improving
accuracy.
Under a strict 10\% token budget, VETO achieves 55.7\%, outperforming
VFlowOpt (54.9\%) and VisionZip (52.6\%).
In the fully supervised setting (Section~\ref{sec:supervised}), VETO
additionally achieves state-of-the-art on long-form narrative benchmarks and competitive procedural action recognition.

\noindent Our contributions are:
\begin{itemize}
  \item \textbf{Hardware-agnostic dual-axis compression:} A hierarchical
    intra-then-inter ordering that resolves the efficiency floor of single-axis
    methods, with a growing advantage on modern optimized attention
    infrastructure.
  \item \textbf{Universal, training-optional plug-in:} VETO integrates into
    diverse frozen VLMs (LLaVA, InternVL, LongVA) with no architectural
    changes, no custom kernels, and no retraining required.
  \item \textbf{Extreme-budget robustness:} VETO establishes a new
    state-of-the-art under severely constrained token budgets (10\%
    retention), outperforming all evaluated single-axis baselines.
\end{itemize}

\section{Related Work}
\label{sec:related_work}

\paragraph{Vision-Language Models.}
VLMs bridge visual perception and natural language
processing~\cite{radford2021learning,li2023blip,liu2023visual,instructblip}.
Contrastive pre-training (CLIP~\cite{radford2021learning}) aligned image and
text at scale; instruction-tuned successors such as
LLaVA~\cite{liu2023visual}, InstructBLIP~\cite{instructblip}, and
Qwen-VL~\cite{bai2023qwenvlversatilevisionlanguagemodel} leveraged large
language models as reasoning engines.
For video, VideoChat~\cite{li2023videochat},
Video-LLaMA~\cite{zhang2023video}, and
Video-ChatGPT~\cite{maaz2023video} extend image-based VLMs to the temporal
domain.
All share the same scalability challenge: processing long videos generates
tens of thousands of visual tokens, creating quadratic attention
complexity that limits practical deployment.

\paragraph{Efficiency plug-ins for VLMs.}
Token-reduction methods include pruning~\cite{rao2021dynamicvit}, merging
(ToMe~\cite{bolya2022token}), and adaptive
selection~\cite{Chen_2023_CVPR}.
FastV~\cite{chen2024image} prunes visual tokens based on attention statistics;
VisionZip~\cite{yang2025visionzip} compresses the visual context for image
VLMs; VFlowOpt~\cite{yang2025vflowopt} guides token pruning with visual
information flow.
ST-LLM~\cite{liu2024st} introduces spatial-temporal token selection for
video.
LLaVA-PruMerge~\cite{shang2024llava} combines pruning and merging for image
VLMs.
VideoLLM-MoD~\cite{wu2024videollm} applies mixture-of-depths to video.
HERMES~\cite{faure2025hermes} provides training-free hierarchical memory for
long videos.
Unlike these works, which operate along a single spatial or temporal axis,
\MethodName~introduces dual-axis compression with a strict hierarchical
ordering—exploiting the complementary nature of spatial and temporal
redundancy to achieve latency reductions that grow with the efficiency of the
underlying attention kernel.

\begin{figure*}[t]
    \centering
    \includegraphics[width=\linewidth]{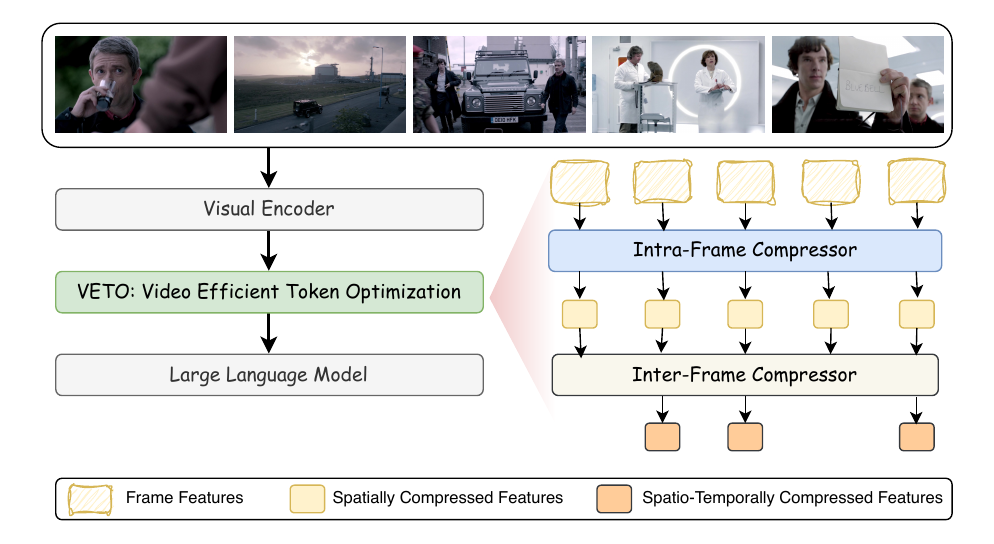}
    \caption{VETO architecture. Visual encoder features are first compressed by
    the \emph{Intra-Frame Compressor} (semantic token merging within each
    frame) and then by the \emph{Inter-Frame Compressor} (frame-level
    merging). The compressed representation maintains feature dimensionality
    and is fed to the frozen LLM without architectural changes.}
    \label{fig:method}
\end{figure*}

\section{VETO: \MethodNameExt}
\label{sec:methodology}

\MethodName~is a training-optional plug-in that addresses the computational
bottleneck in video understanding through dual-axis compression: spatial
compression within frames (\textit{Intra-Frame Compressor}) followed by
temporal compression across the sequence (\textit{Inter-Frame Compressor}).
This hierarchical ordering ensures that the expensive temporal similarity
matching operates on an already-reduced spatial dimension, unlocking
substantial latency reductions.

\subsection{Motivation: The Complexity Bottleneck}
\label{sec:complexity_motivation}

Standard VLM attention scales as $\mathcal{O}((T \cdot N)^2)$.
Single-axis temporal methods reduce $T$ but must first compute a global
similarity matrix across \emph{fully uncompressed} spatial tokens, introducing
a strict hardware-level latency floor that prevents optimal speed-ups.
\MethodName~bypasses this floor through hierarchical ordering.
Intra-frame spatial compression first reduces the spatial dimension to $N/k$
at a cost of only $\mathcal{O}(N^2/k)$ per frame.
The subsequent inter-frame temporal compression then operates on this shrunken
feature space, requiring only $\mathcal{O}(T^2 \cdot C)$ for global similarity
assessment (typically $C \ll N$).
The effective attention complexity becomes
$\mathcal{O}(T \cdot N^2/k + T^2 \cdot C)$—fundamentally breaking the
single-axis latency floor.

\subsection{Problem Formulation}

Given a video $\mathcal{V} = \{I_1, \ldots, I_T\}$, the visual encoder
produces a memory bank $\mathbf{M} \in \mathbb{R}^{B \times T \times N \times C}$.
We seek compression functions
$f_{\text{intra}}: \mathbb{R}^{B \times T \times N \times C} \rightarrow
\mathbb{R}^{B \times T \times N' \times C}$
and
$f_{\text{inter}}: \mathbb{R}^{B \times T \times N' \times C} \rightarrow
\mathbb{R}^{B \times T' \times N' \times C}$
with $N' < N$ and $T' < T$, minimizing information loss:
\begin{equation}
\mathcal{L} = \mathbb{E}_{\mathcal{V}} \left[
  \mathcal{D}\!\left(\mathcal{F}(\mathbf{M}),\,
  \mathcal{F}(f_{\text{inter}}(f_{\text{intra}}(\mathbf{M})))\right)
\right]
\end{equation}
where $\mathcal{F}$ is downstream task performance and $\mathcal{D}$ measures
semantic preservation.

\subsection{Intra-Frame Compressor}
\label{sec:intra-frame}

The Intra-Frame Compressor reduces spatial redundancy within individual frames
by merging semantically similar tokens.
Let $\mathbf{X}_t \in \mathbb{R}^{N \times C}$ be the token features for
frame $t$, extracted from the frozen visual encoder prior to any LLM
projection.

We employ a $k$-way splitting strategy inspired by optimal transport.
Tokens are partitioned into ``destination'' anchors and ``source'' tokens:
\begin{align}
\mathbf{X}_{\text{src}} &= \mathbf{X}_t[\text{idx} \not\equiv 0 \pmod{k},\,:]
  \;\in\mathbb{R}^{N_{\text{src}} \times C} \label{eq:src}\\
\mathbf{X}_{\text{dst}} &= \mathbf{X}_t[\text{idx} \equiv 0 \pmod{k},\,:]
  \;\in\mathbb{R}^{N_{\text{dst}} \times C} \label{eq:dst}
\end{align}
where $N_{\text{dst}} = \lfloor N/k \rfloor$ and $N_{\text{src}} = N -
N_{\text{dst}}$.
We then compute the cosine similarity matrix between source and destination
tokens:
\begin{equation}
\mathbf{S} = \frac{\mathbf{X}_{\text{src}}\,\mathbf{X}_{\text{dst}}^T}
  {\|\mathbf{X}_{\text{src}}\|_2\,\|\mathbf{X}_{\text{dst}}\|_2}
\end{equation}
and assign each source to its most similar destination:
\begin{equation}
\label{eq:argmax}
\text{idx}_i = \arg\max_j \mathbf{S}_{i,j}, \quad i \in [1, N_{\text{src}}]
\end{equation}
Merged tokens are formed by size-weighted averaging:
\begin{equation}
\mathbf{X}'_{t,j} =
  \frac{\sum_{i:\,\text{idx}_i = j} \mathbf{X}_{\text{src},i}
    + \mathbf{X}_{\text{dst},j}}
       {|\{i:\,\text{idx}_i = j\}| + 1}
\end{equation}
This reduces the spatial token count from $N$ to approximately $N/k$, outputting the spatially compressed bank $\mathbf{M}' \in \mathbb{R}^{B \times T \times N' \times C}$.

\noindent\textbf{Rationale for uniform partitioning.}
The modulo-$k$ partition is spatially \emph{uniform by design}: every spatial
region contributes exactly one anchor, guaranteeing complete spatial coverage.
Content-adaptive anchor selection concentrates anchors in salient regions,
leaving peripheral areas without valid destinations—leading to precisely the
spatial collapse one might expect from aggressive compression.
In VETO, content-adaptiveness is reserved for the \emph{merging} step
(Eq.~\ref{eq:argmax}): each source token is matched to its cosine-nearest
anchor, achieving semantic grouping.
Adding content-adaptive anchor scoring would require an additional
$\mathcal{O}(N^2)$ pass, negating the efficiency gain.
Empirically, Table~\ref{tab:ablation_local} confirms accuracy is stable across
$k \in \{2, 3, 4, 5\}$; significant spatial collapse would manifest as
monotonic degradation, which is not observed.

\noindent\textbf{Positional information.}
VETO operates on feature vectors already output by the frozen ViT encoder,
which has baked absolute positional information into each token prior to
compression.
The downstream LLM receives compressed tokens as a flat visual prefix with no
independent spatial position IDs—the standard convention for LLaVA-style
models~\cite{liu2023visual}.
For models with RoPE-based temporal embeddings (e.g.,
LongVA~\cite{zhang2024long}), the LLM assigns sequential indices to the
compressed visual sequence after merging.
Since intra-frame merging groups spatially adjacent tokens (modulo-$k$ groups
span at most $k$ consecutive positions), positional drift is bounded and
empirically harmless, as confirmed by the ablation in
Table~\ref{tab:ablation_local}.

\begin{figure*}[t]
    \centering
    \includegraphics[width=\linewidth]{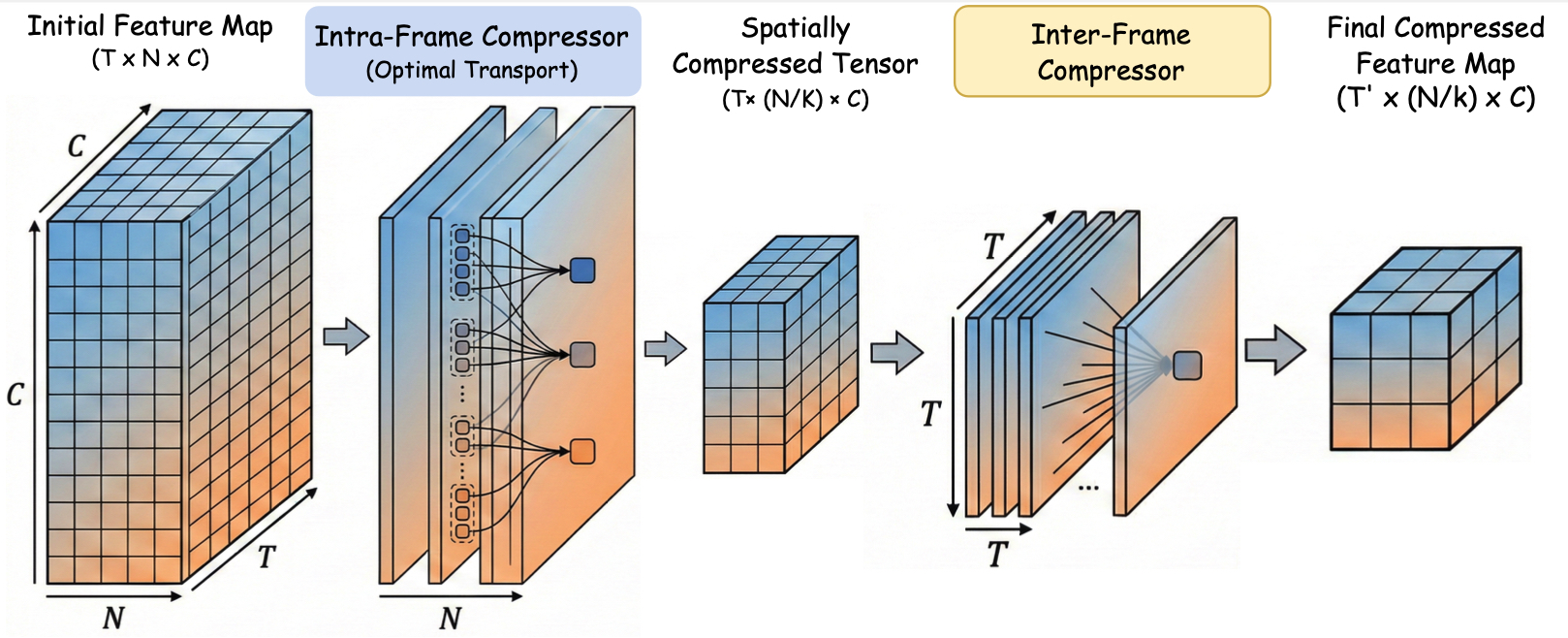}
    \caption{\textbf{The VETO dual-axis pipeline and dimensionality reduction.}
    \textbf{First (Intra-Frame):} Optimal-transport-inspired matching reduces
    spatial token count from $N$ to $N/k$, preserving salient semantics while
    shrinking the feature tensor.
    \textbf{Second (Inter-Frame):} Global frame-level redundancy is eliminated.
    Because the temporal similarity matrix is computed on the already spatially
    compressed dimension ($N/k$), computational overhead is minimized.}
    \label{fig:pipeline_architecture}
\end{figure*}

\subsection{Inter-Frame Compressor}
\label{sec:inter-frame}

The Inter-Frame Compressor addresses temporal redundancy by identifying and
merging similar frames. It operates \emph{after} intra-frame compression, computing
temporal similarity on the already-reduced spatial dimension $N'$.

\subsubsection{Global Similarity Assessment}

Given $\mathbf{M}' \in \mathbb{R}^{B \times T \times N' \times C}$, frame
representations are obtained by spatial mean-pooling:
\begin{equation}
\mathbf{F}_t = \frac{1}{N'} \sum_{n=1}^{N'} \mathbf{M}'_{:,t,n,:}
  \in \mathbb{R}^{B \times C}
\end{equation}
The pairwise cosine similarity matrix is:
\begin{equation}
\mathbf{G}_{i,j} = \frac{\mathbf{F}_i \cdot \mathbf{F}_j}
  {\|\mathbf{F}_i\|_2\,\|\mathbf{F}_j\|_2}, \quad i,j \in [1,T]
\end{equation}
A greedy algorithm then selects the top-$m$ most similar non-overlapping pairs
for merging, applying a valid-pair mask $\mathbf{W}_{i,j} = 1$ iff $i < j$ and
both frames are active.
Selected frame pairs are merged by compression-size weighted averaging:
\begin{equation}
\mathbf{M}''_{j_k} = \frac{\mathbf{M}'_{i_k} \cdot s_{i_k}
  + \mathbf{M}'_{j_k} \cdot s_{j_k}}{s_{i_k} + s_{j_k}}
\end{equation}
This iterates until the target frame count $T'$ is reached.
The number of iterations equals the reduction in frame count (e.g.,
$64 \to 48$ frames requires 16 iterations), not the video length, keeping
the overhead small.

\begin{algorithm}[t]
\caption{Inter-Frame Compression}
\label{alg:inter-frame}
\small
\begin{algorithmic}[1]
    \REQUIRE Memory bank $\mathbf{M}' \in \mathbb{R}^{B \times T \times N' \times C}$, target size $T'$, merges per step $m$
    \ENSURE Compressed memory bank $\mathbf{M}'' \in \mathbb{R}^{B \times T' \times N' \times C}$
    \STATE Initialize active mask $\mathbf{A} \in \{0,1\}^{B \times T}$
    \STATE $\text{count} \gets T$
    \WHILE{$\text{count} > T'$}
        \STATE $\mathbf{F}_t \gets \text{mean}(\mathbf{M}'_{:,t,:,:})$,\quad
               $\mathbf{F}_t \gets \mathbf{F}_t / \|\mathbf{F}_t\|_2$
        \STATE $\mathbf{G} \gets \mathbf{F}\mathbf{F}^T \odot \mathbf{W}$
        \STATE $\{(i_k, j_k)\}_{k=1}^m \gets \text{top-}m(\mathbf{G})$
        \FOR{each pair $(i_k, j_k)$}
            \STATE $\mathbf{M}''_{j_k} \gets \text{weighted\_avg}(\mathbf{M}'_{i_k},\mathbf{M}'_{j_k})$
            \STATE $\mathbf{A}_{i_k} \gets 0$
        \ENDFOR
        \STATE $\text{count} \gets \text{count} - m$
    \ENDWHILE
    \RETURN $\mathbf{M}''$
\end{algorithmic}
\end{algorithm}

\begin{table*}[t]
    \centering
    \small
    \caption{Performance across benchmarks. All latency normalized to Vanilla
    $= 1.00$ ($\downarrow$ is better).
    \textsuperscript{$\dagger$}~Latency on H100+FlashAttention-2.
    \textsuperscript{$\ddagger$}~Latency on V100; full H100 comparison in
    supplementary.
    $\star$~VideoMME accuracy from unified
    re-evaluation~\cite{yang2025vflowopt}: FastV 55.9\%, VisionZip 57.5\%,
    VFlowOpt 58.0\%---all degrade from the 58.9\% Vanilla baseline.}
    \label{tab:main_benchmarks}

    \begin{minipage}[t]{0.57\textwidth}
        \centering
        \setlength{\tabcolsep}{5pt}
        \caption*{(a) \textbf{VideoMME \& LongVideoBench}}
        \begin{tabular}{llcccc}
            \toprule
            \multirow{2}{*}{Model} & \multirow{2}{*}{Method}
              & \multicolumn{2}{c}{VideoMME}
              & \multicolumn{2}{c}{LongVideoBench} \\
            \cmidrule(lr){3-4}\cmidrule(lr){5-6}
              & & Acc. & Lat. & Acc. & Lat. \\
            \midrule
            \multirow{4}{*}{\textit{LLaVA-OneV. (7B)}\textsuperscript{$\dagger$}}
              & Vanilla
                                     & \textbf{58.9} & 1.00 & 57.9 & 1.00 \\
              & ToMe
                                     & 58.3          & 0.96 & 58.3 & 0.98 \\
              & HERMES
                                     & 58.4          & 0.93 & 57.1 & 0.95 \\
              & \cellcolor{myblue}\textbf{VETO (Ours)}
                                     & \cellcolor{myblue}\textbf{58.9}
                                                   & \cellcolor{myblue}\textbf{0.85}
                                                          & \cellcolor{myblue}\textbf{58.4}
                                                                 & \cellcolor{myblue}\textbf{0.86} \\
            \midrule
            \multirow{4}{*}{\textit{LLaVA-OneV. (0.5B)}\textsuperscript{$\ddagger$}}
              & Vanilla
                                     & 42.9          & 1.00 & 46.0          & 1.00 \\
              & ToMe
                                     & \textbf{43.8} & 1.36 & \textbf{46.4} & 0.60 \\
              & HERMES
                                     & 43.5          & 0.92 & 44.8          & 0.61 \\
              & \cellcolor{myblue}\textbf{VETO (Ours)}
                                     & \cellcolor{myblue}43.6
                                                   & \cellcolor{myblue}\textbf{0.91}
                                                          & \cellcolor{myblue}45.9
                                                                 & \cellcolor{myblue}\textbf{0.55} \\
            \midrule
            \multirow{4}{*}{\textit{LongVA (7B)}\textsuperscript{$\dagger$}}
              & Vanilla
                                     & \textbf{54.0} & 1.00 & 52.9 & 1.00 \\
              & ToMe
                                     & 53.3          & 0.93 & 52.7 & 0.95 \\
              & HERMES
                                     & 53.1          & 0.95 & 52.9 & 0.95 \\
              & \cellcolor{myblue}\textbf{VETO (Ours)}
                                     & \cellcolor{myblue}\textbf{53.5}
                                                   & \cellcolor{myblue}\textbf{0.90}
                                                          & \cellcolor{myblue}\textbf{53.0}
                                                                 & \cellcolor{myblue}\textbf{0.86} \\
            \midrule
            \multirow{2}{*}{\textit{InternVL-2.5 (1B)}\textsuperscript{$\ddagger$}}
              & Vanilla
                                     & \textbf{49.2} & 1.00 & \textbf{45.7} & 1.00 \\
              & \cellcolor{myblue}\textbf{VETO (Ours)}
                                     & \cellcolor{myblue}48.1
                                                   & \cellcolor{myblue}\textbf{0.81}
                                                          & \cellcolor{myblue}45.6
                                                                 & \cellcolor{myblue}\textbf{0.75} \\            
            \bottomrule
        \end{tabular}
    \end{minipage}\hfill
    \begin{minipage}[t]{0.40\textwidth}
        \centering
        \setlength{\tabcolsep}{2pt}
        \caption*{(b) \textbf{EgoSchema} (V100 lat.)}
        \begin{tabular}{lcc}
            \toprule
            Method & Acc. & Lat. \\
            \midrule
            \multicolumn{3}{l}{\textit{LLaVA-OneVision (7B)}} \\
            Vanilla & 60.2 & 1.00 \\
            ToMe    & \textbf{60.5} & 0.78 \\
            HERMES  & 60.0 & 0.82 \\
            \rowcolor{myblue}\textbf{VETO} & 60.2 & \textbf{0.60} \\
            \bottomrule
        \end{tabular}

        \vspace{4mm}

        \caption*{(c) \textbf{Image Benchmarks} (LLaVA-OV 7B, V100 lat.)}
        \begin{tabular}{lcc}
            \toprule
            Method & Acc. & Lat. \\
            \midrule
            \multicolumn{3}{l}{\textit{MME}} \\
            Vanilla & \textbf{418/1580} & 1.00 \\
            \rowcolor{myblue}\textbf{VETO} & 418/1571 & \textbf{0.88} \\
            \midrule
            \multicolumn{3}{l}{\textit{MMMU}} \\
            Vanilla & \textbf{48.8} & 1.00 \\
            \rowcolor{myblue}\textbf{VETO} & 48.3 & \textbf{0.73} \\
            \bottomrule
        \end{tabular}
    \end{minipage}
\end{table*}


\section{Experiments}
\label{sec:experiments}

\subsection{Implementation Details}

All plug-in experiments inject \MethodName~after the visual encoder and before
the LLM, preserving per-token feature dimensionality.
Latency is measured end-to-end (visual encoder + token compression + LLM
inference) at batch size~1.
Primary latency results use \textbf{H100+FlashAttention-2}; for
InternVL-2.5-1B and LLaVA-OneVision-0.5B, V100 latency is reported directly
(full H100 comparisons in supplementary material).
Vanilla baseline scores and \MethodName~evaluations share a unified codebase.
Accuracy for FastV~\cite{chen2024image}, VisionZip~\cite{yang2025visionzip},
and VFlowOpt~\cite{yang2025vflowopt} is taken from the unified
re-evaluation in~\cite{yang2025vflowopt}; their hardware latency was not
benchmarked.

Table~\ref{tab:hyperparams} summarizes the hyperparameter settings used
across all benchmarks.
For zero-shot evaluations, VETO is injected without any fine-tuning, using
a fixed intra-frame factor $k=2$ (retaining 50\% of spatial tokens per frame)
and an inter-frame retention ratio $G_{\text{ratio}} = 0.75$ for video
benchmarks.
For fully supervised tasks (Section~\ref{sec:supervised}), only a lightweight
linear adapter is optimized; the visual encoder and LLM remain frozen
throughout training.

\begin{table}[t]
    \centering
    \small
    \setlength{\tabcolsep}{3pt}
    \caption{Hyperparameter settings. $k$: intra-frame factor (retains $1/k$
    spatial tokens per frame); $G$: frame retention fraction.
    \textbf{Abbrevs:}~VME\,=\,VideoMME; LVB\,=\,LongVideoBench;
    ES\,=\,EgoSchema; MC\,=\,MovieChat-1K; BF\,=\,BreakFast.}
    \label{tab:hyperparams}
    \begin{minipage}[t]{0.48\linewidth}
        \centering
        \caption*{(a) Zero-Shot}
        \begin{tabular}{lccc}
            \toprule
             & Frames & $k$ & $G$ \\
            \midrule
            VME & 64 & 2 & 0.75 \\
            LVB & 64 & 2 & 0.75 \\
            ES  & 64 & 2 & 0.75 \\
            MC  & 40 & 2 & 0.50 \\
            \bottomrule
        \end{tabular}
    \end{minipage}\hfill
    \begin{minipage}[t]{0.48\linewidth}
        \centering
        \caption*{(b) Fully Supervised}
        \begin{tabular}{lcccc}
            \toprule
             & Frames & $k$ & $G$ & Ep. \\
            \midrule
            BF   & 100 & 2 & 0.32 & 20 \\
            COIN & 100 & 2 & 0.32 & 20 \\
            \bottomrule
        \end{tabular}
    \end{minipage}
\end{table}

\subsection{Integration with Vision-Language Models}
\label{sec:integration}

\paragraph{Video understanding benchmarks.}
Table~\ref{tab:main_benchmarks}a reports accuracy and normalized latency across
four VLM families.
H100+FlashAttention-2 latency is reported for LLaVA-OV-7B and LongVA-7B
($\dagger$); V100 latency for InternVL-2.5-1B and LLaVA-OV-0.5B ($\ddagger$),
with full H100 results in supplementary material.

On \textbf{LLaVA-OneVision-7B}, VETO achieves \textbf{58.9\%} VideoMME
accuracy (matching Vanilla) with a \textbf{15\% latency reduction}
(0.85$\times$) on H100.
ToMe and HERMES, operating on the \emph{identical} LLaVA-OneVision-7B
backbone, manage only 4\% and 7\% reductions respectively.
VETO's margin is \textbf{3.75$\times$ larger than ToMe's}, not a backbone
advantage, but a structural one: total-sequence-length reduction cannot be
absorbed by fused attention kernels, whereas per-frame spatial reductions can.
As reported in~\cite{yang2025vflowopt}, FastV (55.9\%), VisionZip (57.5\%),
and VFlowOpt (58.0\%) all degrade accuracy below the 58.9\% Vanilla baseline.

\begin{findingbox}{1}
Same backbone, same hardware, same benchmark: VETO cuts latency by
\textbf{15\%} while ToMe manages 4\% and HERMES 7\%.
This \textbf{3.75$\times$ larger margin} comes with \emph{zero accuracy penalty}.
\end{findingbox}

On \textbf{LongVA-7B}, VETO achieves the best accuracy \emph{and} best latency
on both benchmarks simultaneously: 53.5\%/0.90$\times$ on VideoMME and
53.0\%/0.86$\times$ on LongVideoBench, outperforming ToMe (53.3\%/0.93$\times$)
and HERMES (53.1\%/0.95$\times$).
A different architecture, the same story: cross-architecture robustness confirmed.

For \textbf{InternVL-2.5-1B}, ToMe and HERMES lack implementation support;
we compare against the Vanilla baseline.
VETO achieves competitive accuracy (48.1\% vs.\ 49.2\% VideoMME) with a
\textbf{19\% V100 latency reduction}.

\paragraph{Long-term reasoning and image benchmarks.}
On EgoSchema (Table~\ref{tab:main_benchmarks}b), VETO achieves a
\textbf{40\% latency reduction} while matching vanilla accuracy (60.2\%)---nearly
doubling the efficiency gain over ToMe (22\%) and HERMES (18\%), on the
identical LLaVA-OneVision-7B backbone.
For image benchmarks (Table~\ref{tab:main_benchmarks}c), VETO maintains
identical MME perception scores (418/1580) with a 12\% speedup, and achieves
48.3\% on MMMU (vs.\ 48.8\%) with a \textbf{27\% speedup}, confirming that
spatial redundancy is exploited effectively even for static images.

\subsection{Fully Supervised Results}
\label{sec:supervised}

The plug-in experiments establish VETO's deployment value.
But do the same compression principles hold---or amplify---when training is
introduced?
We evaluate a fully-supervised variant where the visual encoder and LLM remain
\emph{strictly frozen}; only a lightweight linear adapter is optimized
(20 epochs; hyperparameters in Table~\ref{tab:hyperparams}b).

\begin{table}[t]
    \centering
    \small
    \setlength{\tabcolsep}{4pt}
    \caption{(a) MovieChat-1K long-form narrative
    understanding (all baselines 7B, zero-shot). (b) Procedural action recognition on
    BreakFast (BF) and COIN (Fully-supervised).
    $\dagger$~Methods using the identical InstructBLIP (with vicuna-7B) backbone as VETO.}
    \label{tab:supervised}
    \begin{minipage}[t]{0.46\linewidth}
        \centering
        \caption*{(a) \textbf{MovieChat-1K}}
        \begin{tabular}{lc}
            \toprule
            Model & Acc. \\
            \midrule
            Video-LLaMA~\cite{zhang2023video}   & 51.7 \\
            VideoChat~\cite{li2023videochat}     & 57.8 \\
            VideoChatGPT~\cite{maaz2023video}    & 47.6 \\
            MovieChat~\cite{song2024moviechat}   & 62.3 \\
            Long-LLaVA~\cite{wang2024longllava}  & 69.6 \\
            TimeChat~\cite{ren2024timechat}      & 73.8 \\
            HERMES$^\dagger$~\cite{faure2025hermes} & 78.6 \\
            \midrule
            \rowcolor{myblue}\textbf{VETO (Ours)} & \textbf{80.6} \\
            \bottomrule
        \end{tabular}
    \end{minipage}\hfill
    \begin{minipage}[t]{0.50\linewidth}
        \centering
        \caption*{(b) \textbf{Procedural Action Recognition}}
        \begin{tabular}{lcc}
            \toprule
            Method & BF & COIN \\
            \midrule
            VIS4mer~\cite{islam2022long}    & 88.2 & 88.4 \\
            TranS4mer~\cite{islam2023efficient} & 90.3 & 89.2 \\
            S5~\cite{wang2023selective}     & 90.7 & 90.8 \\
            MA-LMM$^\dagger$~\cite{he2024ma} & 93.0 & 93.2 \\
            VTM~\cite{lee2024video}          & 91.3 & 88.6 \\
            HERMES$^\dagger$~\cite{faure2025hermes} & \textbf{95.2} & \textbf{93.5} \\
            \midrule
            \rowcolor{myblue}\textbf{VETO (Ours)} & \underline{94.4} & \underline{92.7} \\
            \bottomrule
        \end{tabular}
    \end{minipage}
\end{table}

\paragraph{Long-form narrative understanding.}
Table~\ref{tab:supervised}a shows results on MovieChat-1K.
VETO achieves \textbf{80.6\%} accuracy---a new state-of-the-art and a
\textbf{2.0\%} improvement over HERMES ($\dagger$), which uses the
\emph{identical} InstructBLIP (vicuna-7B) backbone.
This is not a model-size artifact: by eliminating redundant tokens, the frozen
LLM concentrates representational capacity on semantically critical events,
improving coherence across long temporal horizons.

\begin{findingbox}{3}
New SotA on MovieChat-1K: \textbf{VETO 80.6\%} vs.\ HERMES 78.6\%
(same backbone). This shows compression does not just preserve understanding; it enhances it.
\end{findingbox}

\paragraph{Procedural action recognition.}
On BreakFast and COIN (Table~\ref{tab:supervised}b), VETO achieves 94.4\% and
92.7\% respectively, competitive with dedicated temporal architectures and
ranking second overall.
The strong performance across both datasets, despite the compression-focused
design, validates that hierarchical token merging preserves the fine-grained
temporal structure necessary for procedural understanding.

\subsection{Ablation Studies}
\label{sec:ablation}

We conduct ablation studies using LLaVA-OneVision-7B on VideoMME,
LongVideoBench, and MovieChat-1K.

\begin{table*}[t]
\centering
\setlength{\tabcolsep}{4pt}
\caption{Ablations (LLaVA-OV~7B, V100 lat.). (a) Token merging strategy,
revealing the latency floor of single-axis methods. (b) Intra-frame
compression ratios. (c) Inter-frame compression ratios. (d) Token starvation
(VideoMME) vs.\ single-axis baselines.}
\label{tab:ablations}

\begin{minipage}[t]{\textwidth}
    \centering
    \caption*{(a) Merging Strategy \& Latency (LLaVA-OneVision 7B)}
    \begin{tabular}{lccccc}
        \toprule
        \multirow{2}{*}{Strategy} &
          \multicolumn{2}{c}{VideoMME} &
          \multicolumn{2}{c}{LongVideoBench} &
          MovieChat-1K \\
        \cmidrule(lr){2-3}\cmidrule(lr){4-5}
        & Acc. & Lat. & Acc. & Lat. & Acc. \\
        \midrule
        Intra-only    & 58.4 & 1.10 & 58.3 & 1.14 & 80.98 \\
        Inter-only    & 58.4 & 1.16 & 58.2 & 1.16 & 80.59 \\
        Inter-first   & 58.4 & 1.07 & 58.0 & 1.09 & 80.78 \\
        \rowcolor{myblue}\textbf{Intra-first (VETO)} & \textbf{58.9} & \textbf{1.00} & \textbf{58.4} & \textbf{1.00} & \textbf{81.18} \\
        \bottomrule
    \end{tabular}
\end{minipage}

\vspace{3mm}

\begin{minipage}[t]{0.35\textwidth}
    \centering
    \caption*{(b) Intra-Frame Ratio}
    \label{tab:ablation_local}
    \begin{tabular}{lcc}
        \toprule
        Ratio & VideoMME & MovieChat-1K \\
        \midrule
        \rowcolor{myblue}1/2 & \textbf{58.9} & \textbf{80.78} \\
        1/3 & 58.4 & 80.98 \\
        1/4 & 58.4 & 80.78 \\
        1/5 & 59.0 & 80.98 \\
        \bottomrule
    \end{tabular}
\end{minipage}\hfill
\begin{minipage}[t]{0.35\textwidth}
    \centering
    \caption*{(c) Inter-Frame Ratio}
    \label{tab:ablation_global}
    \begin{tabular}{lcc}
        \toprule
        Ratio & VideoMME & MovieChat-1K \\
        \midrule
        \rowcolor{myblue}0.75  & \textbf{58.9} & \textbf{80.78} \\
        0.50  & 57.9 & 79.80 \\
        0.25  & 56.1 & 79.61 \\
        0.125 & 52.1 & 77.65 \\
        \bottomrule
    \end{tabular}
\end{minipage}\hfill
\begin{minipage}[t]{0.25\textwidth}
    \centering
    \caption*{(d) Token Starvation (VideoMME)}
    \label{tab:starvation}
    \begin{tabular}{lcc}
        \toprule
        Method & 25\% & 10\% \\
        \midrule
        FastV     & 55.9 & 47.9 \\
        VisionZip & 57.5 & 52.6 \\
        VFlowOpt  & 58.0 & 54.9 \\
        \rowcolor{myblue}\textbf{VETO}     & \textbf{58.3} & \textbf{55.7} \\
        \bottomrule
    \end{tabular}
\end{minipage}
\end{table*}

\paragraph{Token merging strategy.}
Table~\ref{tab:ablations}a uses Intra-first (VETO) as the reference (1.00)
and measures how all other orderings fare against it.
Every alternative is slower: Intra-only (1.10--1.14$\times$), Inter-only
(1.16$\times$), and even Inter-first (1.07--1.09$\times$)—the seemingly
symmetric counterpart—incur extra overhead because their similarity
computations run on fully uncompressed feature spaces.
Intra-first is uniquely efficient: by shrinking spatial dimensions
\emph{before} temporal matching, the inter-frame similarity step operates on
an already-reduced representation, cutting its cost dramatically.
It also achieves the best accuracy on all three benchmarks simultaneously.

\begin{findingbox}{2}
Every other ordering is \textbf{7--16\% slower} than Intra-first (VETO).
\textbf{Intra-first achieves both best accuracy and lowest latency}.
\end{findingbox}

\paragraph{Effect of intra-frame compression ratio.}
Table~\ref{tab:ablations}b shows remarkable stability: accuracy remains within
0.6\% across $k \in \{2,3,4,5\}$ on VideoMME.
This confirms that individual frames contain substantial spatial redundancy,
and that the uniform-anchor + cosine-nearest-source matching successfully
identifies semantically equivalent tokens without damaging the visual
representation.

\paragraph{Effect of inter-frame compression ratio.}
Table~\ref{tab:ablations}c reveals a clear asymmetry: temporal compression is
far more sensitive than spatial compression.
At $G_{\text{ratio}} = 0.75$, accuracy is fully preserved; below 0.25,
degradation becomes substantial (VideoMME drops from 58.9\% to 52.1\%).
This asymmetry reflects a structural property of video: individual frames
contain redundant textures and backgrounds, but temporal transitions encode
causally critical events that reasoning tasks cannot recover from if lost.

\begin{findingbox}{4}
\textbf{Spatial is forgiving; temporal is not.}
Accuracy degrades $<$0.6\% across 4$\times$ spatial reduction (Table b).
It drops 6.8\% at 4$\times$ temporal reduction (Table c).
Recommended: $k = 2$, $G_{\text{ratio}} \geq 0.75$.
\end{findingbox}

\paragraph{Robustness under extreme token starvation.}
Table~\ref{tab:ablations}d stress-tests all methods at 25\% and 10\% token
retention budgets.
At 10\% retention, VETO achieves \textbf{55.7\%} (the only method above 55\%).
The dual-axis design degrades gracefully as each compressor operates at
a moderate individual ratio rather than forcing one axis to overcompensate.

\begin{findingbox}{5}
At 10\% token budget \textbf{VETO 55.7\%} $>$ VFlowOpt 54.9 $>$
VisionZip 52.6 $>$ FastV 47.9.
Dual-axis compression is uniquely suited to extreme-budget inference.
\end{findingbox}

\subsection{Qualitative Compression Visualizations}

\begin{figure*}[t]
    \centering
    \includegraphics[width=\linewidth]{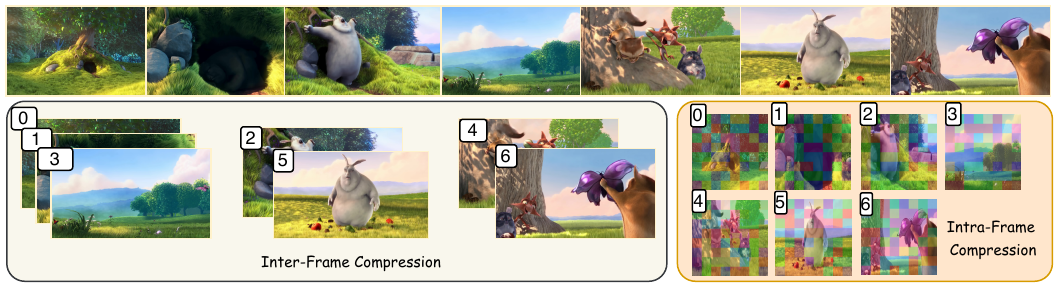}
    \caption{Qualitative visualizations of VETO's dual-axis compression.
    \textbf{Left:} The inter-frame compressor retains scene-transition
    keyframes and discards redundant frames (7 $\to$ 3).
    \textbf{Right:} The intra-frame compressor; patches colored by merge group.
    Semantically similar background regions (sky, grass, textures) are merged
    while distinct foreground objects are preserved.}
    \label{fig:compression_viz}
\end{figure*}

Figure~\ref{fig:compression_viz} visualizes both compression stages.
The inter-frame compressor (left) intelligently retains keyframes that capture
scene transitions, discarding redundant temporal information.
The intra-frame compressor (right) groups semantically similar background
patches while preserving distinct foreground objects—demonstrating that
similarity-based merging preserves semantically critical regions that uniform
downsampling would discard.

\section{Discussion}
\label{sec:discussion}

\paragraph{Spatial vs.\ temporal redundancy asymmetry.}
A recurring theme across our ablations is a fundamental asymmetry between the
two compression axes.
Spatial (intra-frame) compression is highly robust: accuracy on VideoMME
remains within 0.6\% across $k \in \{2,3,4,5\}$, corresponding to retaining
between 20\% and 50\% of per-frame tokens.
Temporal (inter-frame) compression is considerably more sensitive: degradation
becomes substantial below $G_{\text{ratio}} = 0.25$, and moderate at 0.50.
This asymmetry is not accidental—it reflects a structural property of video.
Individual frames encode highly redundant spatial information (textures,
backgrounds, repeated patterns), whereas temporal transitions encode causally
important events and state changes that are critical for reasoning tasks.
VETO's design exploits this asymmetry by applying aggressive spatial
compression first and conservative temporal compression second, leading to
the recommended defaults $k=2$, $G_{\text{ratio}} \geq 0.75$.

\paragraph{Why the advantage grows on modern hardware.}
Fully-fused attention kernels such as FlashAttention-2 reduce the constant
factor of the quadratic attention cost and are especially effective at
absorbing the per-frame spatial reductions that intra-only methods provide.
However, they cannot absorb a reduction in the total sequence length,
which is precisely what VETO's inter-frame stage delivers.
Consequently, as attention infrastructure continues to improve, VETO's margin
over single-axis baselines widens rather than shrinks.
This makes VETO an increasingly compelling choice as inference stacks are
optimized, unlike methods whose advantage is architecture-specific.

\paragraph{Deployment guidelines.}
For practitioners, we recommend the following defaults across deployment
contexts:
(1) \emph{Latency-critical applications} (real-time video QA): $k=2$,
$G_{\text{ratio}}=0.75$ gives 15\% latency reduction on H100+FA2 with no
accuracy loss.
(2) \emph{Budget-constrained edge devices}: $k=4$, $G_{\text{ratio}}=0.50$
is stable spatially but incurs modest temporal accuracy loss—acceptable for
scene-level tasks.
(3) \emph{Extreme resource constraints} (10\% token budget): VETO
outperforms all single-axis baselines (55.7\% vs.\ 54.9\% for VFlowOpt)
because dual-axis compression allows each compressor to operate at a
moderate individual ratio rather than forcing one axis to overcompensate.

\section{Conclusion}
\label{sec:conclusion}

We present VETO, a training-optional dual-axis token compression plug-in for
video VLMs.
As a plug-in, VETO yields consistent latency reductions across model scales
(0.85--0.86$\times$ on LLaVA-OneVision-7B; 0.55$\times$ on
LLaVA-OneVision-0.5B; 0.86--0.90$\times$ on LongVA-7B).
Its advantage over single-axis baselines grows on modern optimized attention
infrastructure, confirming the hardware-agnostic nature of the dual-axis
design principle.
Under extreme token starvation (10\% budget), VETO establishes a new
state-of-the-art, outperforming all single-axis methods by a significant
margin.
In the fully supervised setting (Section~\ref{sec:supervised}), VETO
additionally achieves competitive procedural recognition (Breakfast:
94.4\%, COIN: 92.7\%).

\paragraph{Broader impacts.}
By reducing token counts by up to 50\% spatially and 25\% temporally, VETO
directly reduces FLOPs and energy consumption during inference, contributing
to the goal of sustainable VLM deployment on consumer-grade hardware.

\paragraph{Limitations.}
All plug-in experiments are conducted on models up to 7B parameters.
At larger scales, attention patterns, token redundancy profiles, and hardware
bottlenecks may differ; empirical validation beyond 7B is left to future work.
Additionally, the intra-frame compressor inherits no text-query signal:
compression is purely content-based, not query-aware.
A query-conditioned extension would add an $\mathcal{O}(N^2)$ anchor-scoring
pass but could improve performance on tasks requiring precise spatial
localization.

{
    \small
    \bibliographystyle{ieeenat_fullname}
    \bibliography{main}
}

\clearpage










\section{Complexity Analysis}
\label{sec:supp_complexity}

\noindent\textbf{Notation.} $T$: frames; $N$: tokens per frame; $k$:
intra-frame split factor; $G$: inter-frame retention ratio;
$T' = G \cdot T$; $N' = N/k$; $C$: feature dimension (constant).

\subsection{Vanilla attention cost}

Standard self-attention over a flat sequence of $T \times N$ visual tokens:
\begin{equation}
\mathcal{C}_{\text{vanilla}} = \mathcal{O}\!\left((T \cdot N)^2\right)
= \mathcal{O}(T^2 N^2).
\end{equation}

\subsection{VETO end-to-end cost}

VETO introduces three operations before final attention.

\noindent\textbf{(1) Intra-frame compression} computes a
$N_\text{src} \times N_\text{dst}$ cosine similarity matrix per frame
where $N_\text{src} \approx N(1-1/k)$ and $N_\text{dst} \approx N/k$.
Total over $T$ frames:
\begin{equation}
\mathcal{C}_{\text{intra}} = \mathcal{O}(T \cdot N^2/k).
\end{equation}

\noindent\textbf{(2) Inter-frame similarity} mean-pools $T$ spatially-compressed
frames into $C$-dimensional vectors, then computes a $T \times T$ cosine
matrix:
\begin{equation}
\mathcal{C}_{\text{inter}} = \mathcal{O}(T^2 \cdot C).
\end{equation}
Since $C$ is a fixed LLM hidden dimension while $N$ scales with resolution,
this term is negligible.

\noindent\textbf{(3) Final attention} over the compressed sequence of
$T' \times N'$ tokens:
\begin{equation}
\mathcal{C}_{\text{attn}} = \mathcal{O}((T' \cdot N')^2)
= \mathcal{O}\!\left(\frac{G^2 T^2 N^2}{k^2}\right).
\end{equation}

Total VETO cost (dominant terms):
\begin{equation}
\mathcal{C}_{\text{VETO}}
  = \mathcal{O}\!\left(\frac{TN^2}{k} + T^2 C + \frac{G^2 T^2 N^2}{k^2}\right).
\end{equation}

\subsection{Speedup ratio}

The attention speedup factor is simply:
\begin{equation}
\text{Speedup}_{\text{attn}}
= \frac{T^2 N^2}{\,G^2 T^2 N^2 / k^2\,}
= \left(\frac{k}{G}\right)^{\!2}.
\end{equation}

\noindent\textbf{At default settings} ($k=2$, $G=0.75$):
\begin{equation}
\text{Speedup}_{\text{attn}} = \left(\frac{2}{0.75}\right)^{\!2}
\approx \mathbf{7.1\times}
\end{equation}
reduction in attention FLOPs relative to vanilla.

\subsection{Concrete example at VideoMME settings}

VideoMME uses $T=64$ frames of $N=729$ tokens
($27 \times 27$ ViT patch grid).
Table~\ref{tab:supp_flops} compares operation counts at these settings.

\begin{table}[h]
    \centering
    \small
    \setlength{\tabcolsep}{5pt}
    \caption{Operation counts at $T=64$, $N=729$, $k=2$, $G=0.75$.
    Compression overhead is one to two orders of magnitude smaller than
    the attention saving it enables.}
    \label{tab:supp_flops}
    \begin{tabular}{lcc}
        \toprule
        Operation & Seq.\ length & Rel.\ FLOPs \\
        \midrule
        Vanilla attention   & $46{,}656$ & $1.000\times$ \\
        \midrule
        VETO intra (total)  & ---        & $0.007\times$ \\
        VETO inter          & ---        & ${\ll}0.001\times$ \\
        \rowcolor{myblue}VETO attention & $17{,}520$ & $\mathbf{0.141\times}$ \\
        \bottomrule
    \end{tabular}
\end{table}

VETO's compression overhead (intra + inter) totals $\approx 0.7\%$ of
vanilla attention---negligible.
Net attention FLOPs reduction: $\approx\mathbf{7.1\times}$.

\subsection{Single-axis methods cannot achieve the same reduction}

\textbf{Intra-only} ($k=2$, no temporal compression) yields
$T \times N/k = 64 \times 365 = 23{,}360$ tokens.
Attention FLOPs: $(23{,}360 / 46{,}656)^2 \approx 0.25\times$---a
$4\times$ reduction.

\textbf{Inter-only} ($G=0.75$, no spatial compression) yields
$T' \times N = 48 \times 729 = 34{,}992$ tokens.
Attention FLOPs: $(34{,}992 / 46{,}656)^2 \approx 0.56\times$---a
$1.8\times$ reduction.

VETO's dual-axis achieves $7.1\times \approx 4 \times 1.8$---the two
reductions compound because the two axes are independent.

Additionally, inter-only methods must compute the $T \times T$ similarity
matrix \emph{on full-$N$ spatial vectors}, before any spatial compression
has occurred.
This early large-scale operation introduces a latency floor that the reduced
final attention does not fully offset---which is precisely why Table~2a of
the main paper shows Inter-only latency at $1.16\times$ relative to VETO,
not below 1.00.

\section{InternVL-2.5 on EgoSchema}
\label{sec:supp_egoschema}

Table~\ref{tab:supp_egoschema} extends Table~1b of the main paper
(LLaVA-OV-7B) to InternVL-2.5-1B on EgoSchema.
VETO achieves a \textbf{23\% latency reduction} with only a 0.7\%
accuracy drop---confirming cross-architecture efficiency.
ToMe and HERMES are excluded due to limited InternVL-2.5 support.

\begin{table}[h]
    \centering
    \small
    \setlength{\tabcolsep}{5pt}
    \caption{EgoSchema on InternVL-2.5-1B (V100 lat.).
    Accuracy (\%) and normalized latency ($\downarrow$ better).}
    \label{tab:supp_egoschema}
    \begin{tabular}{lcc}
        \toprule
        Method & Acc. & Lat. \\
        \midrule
        \multicolumn{3}{l}{\textit{InternVL-2.5 (1B)}} \\
        Vanilla                               & \textbf{39.8} & 1.00 \\
        \rowcolor{myblue}\textbf{VETO (Ours)} & 39.1          & \textbf{0.77} \\
        \bottomrule
    \end{tabular}
\end{table}

Combined with Table~1b (LLaVA-OV-7B: 60.2\%/0.60$\times$), VETO achieves
consistent latency gains across two architecturally distinct families on
this benchmark.

\section{Standard Action Recognition: UCF-101}
\label{sec:supp_ucf}

Table~\ref{tab:supp_ucf101} evaluates the fully-supervised variant from
the main paper on UCF-101---a short-clip benchmark with \emph{less}
temporal redundancy than the long-form tasks in the main paper.

\begin{table}[h]
    \centering
    \small
    \setlength{\tabcolsep}{5pt}
    \caption{UCF-101 action recognition. VETO uses the same frozen-backbone
    / linear-adapter setup as the main paper's supervised experiments.}
    \label{tab:supp_ucf101}
    \begin{tabular}{lc}
        \toprule
        Method & Acc.~(\%) \\
        \midrule
        TS ConvNets~\cite{simonyan2014two} & 88.0 \\
        C3D~\cite{tran2015learning}        & 82.3 \\
        VideoMAE~\cite{tong2022videomae}   & 91.3 \\
        TSN~\cite{wang2016temporal}        & 94.2 \\
        TSM~\cite{lin2019tsm}              & \textbf{95.9} \\
        \midrule
        \rowcolor{myblue}\textbf{VETO (Ours)} & \underline{94.4} \\
        \bottomrule
    \end{tabular}
\end{table}

\paragraph{Training setup.}
100 frames per clip, $k=2$, $G_{\text{ratio}}=0.20$ (aggressive temporal
compression reflects short clips having less inter-frame redundancy than
long instructional video), 20 epochs, batch size~4.

\paragraph{Interpretation.}
VETO (94.4\%) matches TSN (94.2\%) and falls within 1.5\% of TSM (95.9\%),
a specialist two-stream architecture optimized specifically for action
recognition.
The result demonstrates that the uniform-anchor spatial matching preserves
discriminative features even at $k=2$ compression, and that the dual-axis
design generalizes to temporal scales well outside its primary design target.

\section{Extended Ablation Discussion}
\label{sec:supp_ablation}

This section provides deeper mechanistic analysis of the ablation results
in Table~2 of the main paper.

\paragraph{Why does Intra-only inflate latency?}
Intra-only compresses each frame from $N$ to $N/k$ tokens but applies
no temporal compression.
At default settings ($T=64$, $k=2$): final sequence has
$64 \times 365 = 23{,}360$ tokens.
VETO's final sequence has $48 \times 365 = 17{,}520$ tokens.
The Intra-only sequence is $1.33\times$ longer, producing the $1.10\times$--$1.14\times$
latency overhead seen in Table~2a.

\paragraph{Why does Inter-only inflate latency more than Intra-only?}
Inter-only retains all $N=729$ spatial tokens per frame, yielding
$48 \times 729 = 34{,}992$ tokens---twice as many as VETO.
Worse, the inter-frame similarity step runs on full-$N$ feature vectors,
adding expensive overhead \emph{before} any token reduction is realized.
Both effects compound to produce $1.16\times$ latency---the highest of all
four configurations.

\paragraph{Why does Inter-first still incur overhead despite reaching the
same final token count as Intra-first?}
Both Inter-first and Intra-first end with $T' \times N/k$ tokens.
The difference is entirely in the inter-frame similarity step:
\begin{itemize}
  \item \textbf{Intra-first}: inter similarity computed on $N/k=365$-dim
    mean-pooled vectors $\Rightarrow$ cheaper.
  \item \textbf{Inter-first}: inter similarity computed on $N=729$-dim
    mean-pooled vectors $\Rightarrow$ 2$\times$ more expensive.
\end{itemize}
This single upstream difference accounts for the $1.07\times$--$1.09\times$
overhead of Inter-first.
The lesson: it is not enough to achieve the same \emph{final} compression
ratio---the \emph{order} of operations determines the overhead incurred
along the way.

\paragraph{Spatial vs.\ temporal sensitivity: the asymmetry explained.}
The intra-frame ablation (Table~2b) shows $<$0.6\% VideoMME degradation
across $k \in \{2,3,4,5\}$ (retaining 20--50\% of per-frame tokens).
The inter-frame ablation (Table~2c) shows 6.8\% degradation at
$G_{\text{ratio}} = 0.125$ (retaining 12.5\% of frames).

The structural explanation: a frame's $N$ tokens encode a static scene
where neighboring patches are often semantically redundant (uniform
backgrounds, repeated textures).
The uniform modulo-$k$ partition guarantees every spatial region
contributes one anchor, so merging 50--80\% of tokens is safe.

Temporal frames, by contrast, encode a causal sequence of events.
Each discarded frame may contain a unique transition or action state
that cannot be inferred from neighbors.
This is why degradation becomes non-linear below $G_{\text{ratio}} = 0.50$:
the model begins discarding irreplaceable causal information.

The recommended operating point ($k=2$, $G_{\text{ratio}} = 0.75$) sits
in the robust regime of both axes, validated empirically by Table~2b--c.



\end{document}